\documentclass[runningheads]{llncs}
\usepackage{graphicx}
\usepackage{float}

\begin{document}
\title{Scalable Educational Question Generation with Pre-trained Language Models}
\titlerunning{Scalable Educational Question Generation with PLMs }
% \author{Sahan Bulathwela\inst{1}\orcidID{0000-1111-2222-3333} \and

\author{Sahan Bulathwela, Hamze Muse and Emine Yilmaz
}

\institute{Centre for Artificial Intelligence,
University College London, United Kingdom\\
\email{\{m.bulathwela, hamze.muse.20, emine.yilmaz\}@ucl.ac.uk}
}
\maketitle              % typeset the header of the contribution
\begin{abstract}
% With the boom of digital educational materials and scalable e-learning systems, the potential for realising AI-assisted personalised learning has skyrocketed. In this landscape, the 
The automatic generation of educational questions will play a key role in scaling online education, enabling self-assessment at scale when a global population is manoeuvring their personalised learning journeys. We develop \textit{EduQG}, a novel educational question generation model built by adapting a large language model. Our extensive experiments demonstrate that \textit{EduQG} can produce superior educational questions by further pre-training and fine-tuning a pre-trained language model on the scientific text and science question data. 
% \keywords{First keyword  \and Second keyword \and Another keyword.}
\end{abstract}

\section{Introduction}
Digital learning resources such as Massively Open Online Courses (MOOC) and Open Educational Resources (OER) are abundant, but they often lack associated questions that enable self-testing and skill verification \cite{truelearn,bulathwela2022sus,semantic_truelearn} once the learning resources are consumed. Generating scalable educational questions is crucial for democratising education \cite{ai_ed_demo}. While existing language models are used for question generation, their utility in education has only been explored recently. This work demonstrates how a large language model can be adapted for educational question generation. The experiments validate the improvement of questions through additional pre-training with educational text. The study also explores the impact of pre-training data size on question generation and investigates the enhancement of educational questions through fine-tuning with a science question dataset. The experimental results show that pre-training and fine-tuning with domain-specific scientific text can outperform a   state-of-the-art baseline, providing significant evidence for building an effective educational question-generation model.

\section{Related Work} \label{sec:lit}
%%%%%%%%%%%%%%%%%%%%%%%%%%%%%%%%%%%%%%%%%%%%%%%%%%%%%%%%%%%%%%%%%%%%%%%%%%%%%%%%%%%%%%%%
% This work aims to be a foundational step towards creating AI systems that can generate educational questions for technology-enhanced learning systems. i) Question Generation (QG), where a model is trained to generate a question based on an information extract and ii) Question Answering (QA), where a model is trained to generate a response to a question. QG is a pre-requisite for question answering  are two main sub-tasks that fall under reading comprehension tasks. 
%%%%%%%%%%%%%%%%%%%%%%%%%%%%%%%%%%%%%%%%%%%%%%%%%%%%%%%%%%%%%%%%%%%%%%%%%%%%%%%%%%%%%%%%%%%

This work focuses on developing AI systems capable of generating educational questions for technology-enhanced learning. It involves two main sub-tasks: Question Generation (QG), where a model generates a question based on given information, and Question Answering (QA), where a model generates a response to a question. QG is essential for QA and both tasks are part of reading comprehension tasks. This paper focuses on QG specifically.

% In both scenarios, a text extract that provides the background information is available, we call it the \emph{context}.
% QG can be thought of as a pre-requisite to QA as a question needs to be already available for a QA system to answer it although the context text would be readily available in both the scenarios.  

\subsection{Automatic Question Generation (QG)}

Automatic question generation involves creating valid and coherent questions based on given sentences and desired responses. Previous approaches have used rule-based and neural-based models, with neural models dominating in various applications \cite{zhang2021review}. Recent advancements in deep learning have led to the adoption of sequence-to-sequence models. By leveraging question-answering datasets, neural models can generate questions using both the context and expected response, ensuring high-quality questions. However, this approach often relies on an additional system to identify relevant responses \cite{tamang2022automatic}, limiting its real-world applicability. The scarcity of public datasets also hinders the development of QG systems that generate both questions and answers. Alternatively, QG models can be trained to rely solely on the context, allowing the creation of questions that belong to a specific type \cite{wu2020question} for the document, paragraph, or sentence level \cite{guo2018soft,du2017learning}. This work specifically focuses on the latter task setting, where only the context is used as input.

\subsection{Pre-trained Language Models (PLMs) for Educational QG}

In the field of educational neural question generation, state-of-the-art (SOTA) systems leverage pre-trained language models (PLMs) such as GPT-3 \cite{brown2020language} and Google T5 \cite{raffel2020exploring}. These models, pre-trained on massive text corpora, enable zero-shot question generation without additional training. Recent research has demonstrated the potential for generating educational questions using GPT models \cite{wang2022towards,bhat2022towards}.

Leaf, a cutting-edge question generation system, fine-tunes a large language model for the question and multiple-choice distracter generation \cite{lopez2021simplifying}. It uses the SQuAD 1.1 dataset \cite{DBLP:rajpurkar2016squad} to train its question generation component by fine-tuning a pre-trained T5 model \cite{raffel2020exploring}. This work diverges from SOTA approaches by employing pre-training to further enhance the PLM's handling of scientific language in the educational context \cite{muse2023pre}, a technique that has shown promise in domain-specific applications like medicine \cite{https://doi.org/10.48550/arxiv.2109.04588}.

Our hypothesis is that pre-training with scientific text can lead to better educational question generation even when models are fine-tuned for general-purpose tasks. To evaluate the quality of generated questions, various metrics are utilized, such as BLEU, ROUGE, METEOR, F1-Score, Human Ratings, Perplexity, and Diversity \cite{bhat2022towards,wang2022towards,lopez2021simplifying}. This study selects a representative subset of these metrics to measure success in terms of linguistic validity and fluency.  

\subsection{Related Datasets} \label{sec:datasets}
S2ORC is a corpus comprising 81.1 million English scholarly publications across various academic fields \cite{lo-etal-2020-s2orc}. For question generation (QG) and question-answering (QA) datasets, \cite{zhang2021review} offers a comprehensive review. The Leaf system, our baseline, is designed for educational purposes by fine-tuning the T5 model using the SQuAD 1.1 dataset, which focuses on reading comprehension \cite{DBLP:rajpurkar2016squad}. However, this dataset is less suited for evaluating educational QG capabilities.

In contrast, SciQ \cite{welbl-etal-2017-crowdsourcing} is a collection of 13,679 crowd-sourced scientific exam questions covering physics, chemistry, and other sciences. Although smaller than SQuAD, SciQ is more relevant for objectively evaluating educational QG models. Therefore, we use the SciQ dataset to assess the models developed in this work, aligning our evaluation with real-world scenarios.
 
% While large language models capture a lot of information about the world \cite{raffel2020exploring}, these models need to be pre-trained further in domain-specific datasets to improve their knowledge and fluency in specific domains (e.g. medicine \cite{https://doi.org/10.48550/arxiv.2109.04588}). 

\section{Methodology}

This study aims to study the effect of further pre-training and fine-tuning the Pre-trained Language Model (PLM) on Educational QG.
% The primary objective of our study is to validate if further pre-training and fine-tuning the Pre-trained Language Model (PLM) on educational data can improve educational QG. 
% This work allows us to assess the feasibility of using PLMs to create high-quality educational questions. 

\subsection{Research Questions} \label{sec:rq}
% Our methodology aims to address several research questions. 
\begin{itemize}
    \item \textbf{RQ1:} Can PLMs generate human-like educational questions?
    \item \textbf{RQ2:} Does pre-training PLMs with scientific text improve educational QG?
    \item \textbf{RQ3:} How does the training dataset size affect the pre-training?
    \item \textbf{RQ4:} Does fine-tuning the model with educational questions improve it?
\end{itemize}

\subsection{Question Generations Models} \label{sec:models}

% The research questions outlined in this section 
Our experiments develop QG systems that utilise different PLMs  trained using different task settings. It is important to note that we were not interested in training a neural model from scratch as this is impractical in real-world scenarios due to data scarcity and computational cost \cite{brown2020language}. Instead, we used a PLM as the foundation of the different QG systems we developed for our experiments. 

\subsubsection{Baseline Leaf Model: }
Based on the relevant literature, we identified Leaf system \cite{vachev2022leaf} as the state-of-the-art educational question generation system to use as our baseline. In Leaf, the pre-trained language model, T5, a text-to-text transformer-based language model \cite{raffel2020exploring} (already trained on web-crawled data and Wikipedia articles) is fine-tuned for question generation using a reading comprehension dataset.  

\subsubsection{Proposed EduQG Models:} 
The key differentiator between the baseline model and our proposal is that the EduQG model uses an additional pre-training step that further trains the PLM with scientific text documents before fine-tuning it for question generation. The expectation here is that the additional pre-training on scientific text is going to provide the PLM with more understanding of scientific language and knowledge that is relevant for generating good educational questions. 

We also develop \emph{Leaf+} and \emph{EduQG+}, extending the Leaf model and the EduQG model that is further fine-tuned using an educational question dataset that is more specialised than a reading comprehension dataset that only contains general-purpose questions. We hypothesise that further pre-training harnesses the model's ability to generate educational questions. 

\subsection{Data}
There are different types of datasets that are utilised in different stages of training the PLMs unto  question generation models. These datasets allow us to:
\begin{enumerate}
    \item Pre-train the PLM further with additional scientific language data
    \item Fine-tune the PLM to carry out question generation, which is different from the initial task it was trained on
    \item Objectively evaluate the performance of the question generation model
\end{enumerate}

We incorporate a subset of datasets described in section \ref{sec:datasets} in our experiments. While the pre-training step is skipped when building the baseline Leaf model, the S2ORC corpus \cite{lo-etal-2020-s2orc} is used for pre-training the EduQG models. The resultant language model is fine-tuned for question generation using the SQuAD 1.1 dataset \cite{DBLP:rajpurkar2016squad}. Finally, we use the test set data from the SciQ question dataset \cite{welbl-etal-2017-crowdsourcing} for evaluation. This is because the SciQ dataset exclusively contains science questions from examinations making it suitable for objectively evaluating the model's suitability in \emph{educational question generation}.

\subsection{Evaluation Metrics}

 % The two settings lead to the baseline (Leaf) and the proposed models (EduQG) that we compare.
 As identified in section \ref{sec:lit}, two aspects of quality are considered when evaluating the QG models, i) the prediction accuracy  and ii) the linguistic quality of the generated questions. To measure the predictive accuracy of the questions, we use the BLEU score and the F1 score that is used in prior work \cite{DBLP:rajpurkar2016squad,bhat2022towards,lopez2021simplifying}. To measure how human-like the generated questions are (i.e. linguistic quality), we use perplexity and diversity \cite{wang2022towards}. A lower perplexity score indicates better coherence. The diversity score indicates how diverse the vocabulary of the generated questions is. Larger diversity values coupled with low perplexity, indicate the use of a richer vocabulary with grammatical precision.

\subsection{Experimental Setup} \label{sec:exp}

Our experiments are designed to answer the research questions that are outlined in section \ref{sec:rq}. 
% The goal of RQ1 is to verify if machine-generated questions possess good enough linguistic quality. 
To address RQ1, we calculate the linguistic quality-related metrics (specifically, perplexity and diversity) of the human-generated questions (the ground truth) in the SQuAD 1.1 and SciQ datasets. We hypothesise that the machine-generated questions are acceptable if they demonstrate superior or similar linguistic quality metrics in comparison to the metrics computed using the human-generated questions in the datasets (SQuAD and SciQ).  
The source code is available publicly \footnote{\url{https://github.com/hmuus01/Educational\_QG}}.

\figurename { \ref{fig:method} illustrates the experiments we set out to answer RQs 2-4. The foundational language model to all the developed models (baselines and proposals) is the \emph{T5-small} language model (hereafter referred to as T5 model). Altogether 5 models are developed (coloured boxes in the figure), all of which are evaluated using the SciQ test data.
% The experiment we set up to answer RQ2 is illustrated in \. 
 \begin{figure}[]
    \centering
    \includegraphics[width=\columnwidth]{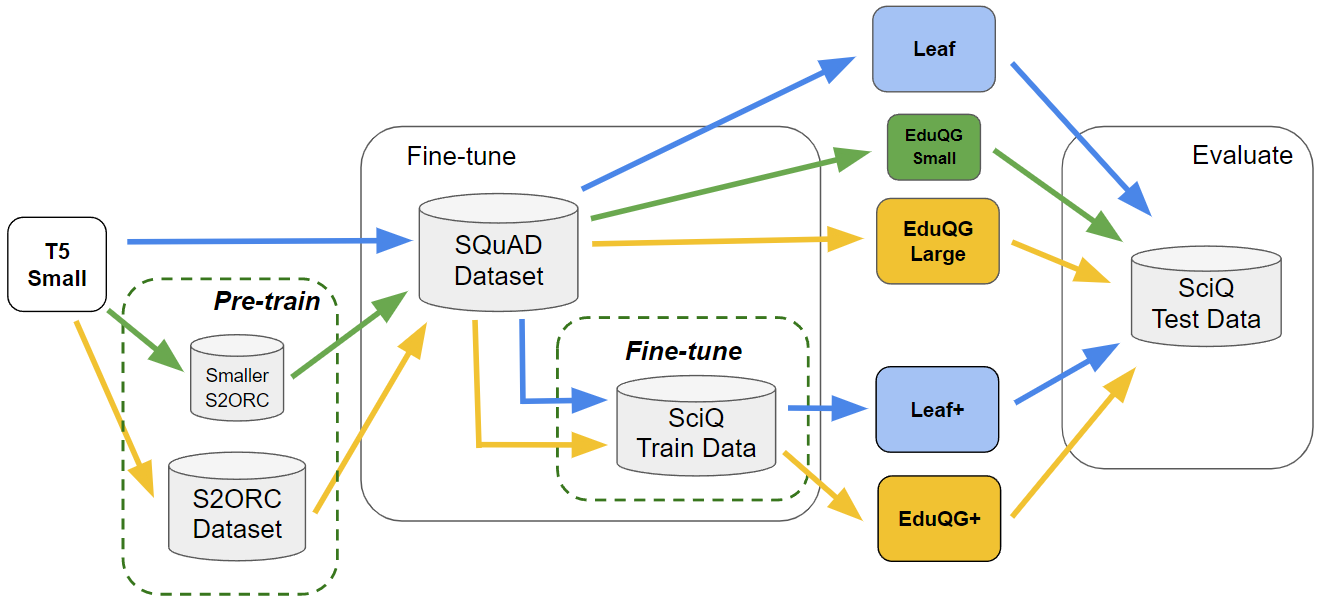}
    \caption{Methodology for training and evaluating the baseline Leaf model (blue),  novel EduQG Small (green) and EduQG Large (yellow) models (and their $\cdot \ +$ counterparts), introducing additional pre-training and fine-tuning steps (green dashed boxes) to address RQ 2,3 and 4.}
    \label{fig:method}
\end{figure}

 To address RQ2, we develop \emph{Leaf} and \emph{EduQG Large} models as per \figurename{ \ref{fig:method}}. As the baseline, we develop the \emph{Leaf} model by fine-tuning the T5 model on the SQuAD 1.1 dataset
% as our baseline QG system 
(blue flow of arrows in \figurename { \ref{fig:method} through the \emph{Leaf} model). Our proposal, \emph{EduQG Large}, additionally pre-trains the T5 model with a down-sampled version of the S2ORC dataset that contains approx. {23.2M} scientific abstracts related to Chemistry, Biology and Physics research papers (yellow flow of in \figurename { \ref{fig:method} through the \emph{EduQG Large} model).  
To answer RQ3, we use two models, i) \emph{EduQG Large} from the previous experiment, and ii) \emph{EduQG Small} (green flow of arrows through the \emph{EduQG Small} model) using a smaller number of training examples from 23.5M data points. 
To answer RQ4, we develop \emph{Leaf+} and \emph{EduQG+} (blue and yellow flows of arrows passing through the $\cdot \ +$ models), extensions of the Leaf and EduQG Large models (baselines for RQ4 experiment) that are further fine-tuned using the training data from the SciQ dataset. While the SQuAD dataset will help the PLM to learn question generation in general, the SciQ training data is expected to teach the model \emph{educational question generation}. We hypothesise this change will lead to superior performance.

% \begin{figure}[]
%     \centering
%     % \includegraphics[width=70mm,height=39mm]{Distribution_Train Data LR.png}
%     % \includegraphics[width=95mm, height=65mm]{LaTeX/methodology_new.png}
%     \includegraphics[width=\columnwidth]{rq4_method.png}
%     \caption{Methodology for training and evaluating Leaf model (blue),  and EduQG (yellow) models with additional pre-training and fine-tuning steps, leading to new models (green dashed boxes).}
%     \label{fig:method_plus}
% \end{figure}

% \paragraph{Evaluation} 
% The two settings lead to the baseline (Leaf) and the proposed model (EduQG) that we compare using the SciQ dataset, as it contains exclusively educational questions.
% To measure the predictive power of the human-generated questions, we use the BLUE score and the F1 score \cite{DBLP:rajpurkar2016squad}. To measure how human-like the generated questions are, we use perplexity, diversity and grammatical error rates. A lower perplexity score indicates better coherence \cite{wang2022towards}.

\section{Results} \label{sec:results}

As per section \ref{sec:exp}, several experiments are executed.
Table \ref{tab:human} shows the perplexity and diversity scores computed on the human-generated questions found in SQuAD 1.1 and SciQ datasets (RQ1). Table \ref{tab:results} presents the prediction accuracy and the linguistic quality metrics calculated for the models described in section \ref{sec:models} (RQ 2 and 3). \figurename{ \ref{fig:metrics}} further elaborates the distribution of metric scores across the test data. Table \ref{tab:resultsp} presents the improvement of predictive performance and the linguistic quality of the models \emph{Leaf+} and \emph{EduQG+} which are further fine-tuned using the SciQ training data (RQ4). Finally, Table {\ref{fig:examples}} shows a handful of randomly selected test examples from the SciQ dataset where the baseline \emph{Leaf} and the novel \emph{EduQG Large} models have generated questions using the same context.

% \begin{table}[] \centering \
% % small
% \begin{tabular}{c|ccccc|ccc}
% \hline
%                 & \multicolumn{5}{c}{Predictive Performance}                          & \multicolumn{3}{c}{Linguistic Quality}       \\
% Model           & BLEU-1 $\uparrow$     & BLEU-2 $\uparrow$     & BLEU-3 $\uparrow$     & BLEU-4   $\uparrow$   & F1-Score  $\uparrow$       & Perplexity $\downarrow$       & Diversity $\uparrow$       & Grammar Errors $\downarrow$       \\
% % &&&&&&&&\\

% \hline
% Leaf (Baseline) & 27.07          & 20.22          & 17.17          & {16.46} & 30.90          & \textbf{30.82} & 0.735          & \textbf{0.102} \\
% EduQG (Ours)        & \textbf{29.19} & \textbf{21.69} & \textbf{18.03} & \textbf{16.76} & \textbf{33.18} & 34.36          & \textbf{0.749} & 0.122  \\       
% \hline
% \end{tabular}
% \caption{Comparison of predictive performance and linguistic quality between Leaf (baseline) and EduQG (our proposal). The superior performance is indicated in \textbf{bold} face.}
% \label{results}
% \end{table}

\begin{table}[] \centering 
\caption{Linguistic quality of the human-generated questions in the datasets.}\label{tab:human}
\begin{tabular}{|c|c|c|}
\hline
                % & \multicolumn{5}{c}{Predictive Performance}                          & \multicolumn{3}{c}{Linguistic Quality}       \\
{Dataset}           & Perplexity $\downarrow$       & Diversity $\uparrow$  \\
% &&&&&&&&\\
\hline
SQuAD 1.1 &  84.16 & 0.779 \\
SciQ & 18.74 & 0.824 \\
\hline
\end{tabular}
\end{table}

\begin{table}[] \centering \small
\caption{Comparison of predictive performance and linguistic quality between Leaf (baseline) and EduQG (our proposals). The best and second best performance is indicated in \textbf{bold} and \emph{italic} faces respectively. The proposed models that outperform the baseline counterpart ($p< 0.01$ in a one-tailed paired t-test) are marked with $\cdot^{(*)}$.}\label{tab:results}
\begin{tabular}{|c|l|l|l|l|l|l|l|}
\hline
&\multicolumn{5}{c|}{Predictive Performance}  & \multicolumn{2}{c|}{Linguistic Quality}       \\
                % & \multicolumn{5}{c}{Predictive Performance}                          & \multicolumn{3}{c}{Linguistic Quality}       \\#
{Model}           & BLEU-1 $\uparrow$    & BLEU-2 $\uparrow$   & BLEU-3 $\uparrow$  & BLEU-4 $\uparrow$  & F1-Score $\uparrow$  & Perplexity $\downarrow$       & Diversity $\uparrow$ \\
% &&&&&&&&\\

\hline
% \emph{Baseline} &&&&&&&\\
Leaf & 27.07          & 20.22          & 17.17          & \emph{16.46} & 30.90  & \textbf{30.82} & 0.735   \\
\hline
% \emph{Our Proposals} &&&&&&&\\
EduQG Small        & \emph{29.07}$^{(*)}$ & \emph{21.52}$^{(*)}$ & \emph{17.49}$^{(*)}$ & {15.94} & \emph{33.12}$^{(*)}$  & 34.51 & \emph{0.736} \\ 
EduQG Large        & \textbf{29.19}$^{(*)}$ & \textbf{21.69}$^{(*)}$ & \textbf{18.03}$^{(*)}$ & \textbf{16.76}$^{(*)}$ & \textbf{33.18}$^{(*)}$  & \emph{34.36}  & \textbf{0.749}$^{(*)}$ \\
\hline
\end{tabular}
\end{table}

\begin{figure}[] \centering
\includegraphics[width=.9\textwidth]{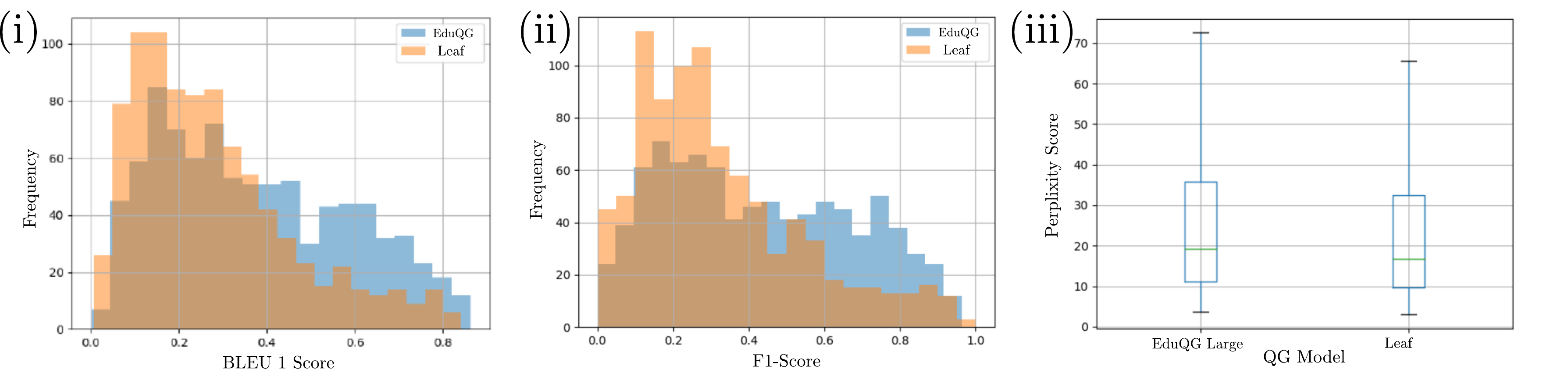}
\caption{The distribution of (i) BLEU 1, (ii) F1 and (iii) Perplexity Score between the Leaf and EduQG models.} \label{fig:metrics}
\end{figure}

\begin{table}[H]
\centering
\scriptsize % Use smaller font size
\caption{Randomly selected contexts From SciQ test data used to create questions using the Leaf and EduQG Large models.} 
\begin{tabular}{|p{0.45\linewidth}|p{0.27\linewidth}|p{0.27\linewidth}|}
\hline
\multicolumn{1}{|c|}{\textbf{Context}} & \multicolumn{1}{c|}{\textbf{EduQG}}  & \multicolumn{1}{c|}{\textbf{Leaf}}                                                  \tabularnewline \hline
% Your table content here
(1) Scientific models are useful tools for scientists. Most of Earth's systems are extremely complex. Models allow scientists to work with systems that are nearly impossible to study as a whole. Models help scientists to understand these systems. They can analyze and make predictions about them using the models. There are different types of models.                                             & What is used to analyze and make predictions about systems that are nearly impossible or easy to study as a whole? & What help scientists to understand the systems of Earth?

\tabularnewline \hline

 (2) Muscles That Move the Head The head, attached to the top of the vertebral column, is balanced, moved, and rotated by the neck muscles (Table 11.5). When these muscles act unilaterally, the head rotates. When they contract bilaterally, the head flexes or extends. The major muscle that laterally flexes and rotates the head is the sternocleidomastoid. In addition, both muscles working together are the flexors of the head. Place your fingers on both sides of the neck and turn your head to the left and to the right. You will feel the movement originate there. This muscle divides the neck into anterior and posterior triangles when viewed from the side (Figure 11.14). & What is the major muscle that laterally rotates? &  What is the major muscle that laterally flexes and rotates the head?
                                            
\tabularnewline \hline
(3) Biodiversity refers to the variety of life and its processes, including the variety of living organisms, the genetic differences among them, and the communities and ecosystems in which they occur. Scientists have identified about 1.9 million species alive today. They are divided into the six kingdoms of life shown in the Figure below. Scientists are still discovering new species. Thus, they do not know for sure how many species really exist today. Most estimates range from 5 to 30 million species. & What term refers to the variety of life and its processes? & How many species are identified today?

\tabularnewline \hline
(4) Take-Home Experiment: The Pupil Look at the central transparent area of someone\u2019s eye, the pupil, in normal room light. Estimate the diameter of the pupil. Now turn off the lights and darken the room. After a few minutes turn on the lights and promptly estimate the diameter of the pupil. What happens to the pupil as the eye adjusts to the room light? Explain your observations. The eye can detect an impressive amount of detail, considering how small the image is on the retina. To get some idea of how small the image can be, consider the following example. &  What is the central transparent area of someone‘s eye? & What is the name of a take-home Experiment?

\tabularnewline \hline 
(5) In both eukaryotes and prokaryotes, ribosomes are the non-membrane bound organelles where proteins are made. Ribosomes are like the machines in the factory that produce the factory's main product. Proteins are the main product of the cell. & What are the non-membrane bound organelles where proteins are made? &  What is the main product of a cell?

\tabularnewline \hline
\end{tabular}
% \caption{Randomly selected contexts From SciQ test data used to create questions using the Leaf and EduQG Large models.} 
\label{fig:examples}
\end{table}
\vspace{-9mm}

\begin{table}[] \centering \small
\caption{Comparison of predictive performance and linguistic quality between Leaf and EduQG models in Table \ref{tab:results} to the new proposals further fine-tuned on SciQ training data, \emph{Leaf+} and \emph{EduQG+}. The best and second best performance is indicated in \textbf{bold} and \emph{italic} faces respectively. The new models that outperform the baseline counterparts ($p< 0.01$ in a one-tailed paired t-test) are marked with $\cdot^{(*)}$.}\label{tab:resultsp}
\begin{tabular}{|c|l|l|l|l|l|l|l|}
\hline
&\multicolumn{5}{c|}{Predictive Performance}  & \multicolumn{2}{c|}{Linguistic Quality}       \\
                % & \multicolumn{5}{c}{Predictive Performance}                          & \multicolumn{3}{c}{Linguistic Quality}       \\#
{Model}           & BLEU-1 $\uparrow$    & BLEU-2 $\uparrow$   & BLEU-3 $\uparrow$  & BLEU-4 $\uparrow$  & F1-Score $\uparrow$  & Perplexity $\downarrow$       & Diversity $\uparrow$ \\
% &&&&&&&&\\

\hline
% \emph{Baseline} &&&&&&&\\
Leaf         & 27.07          & 20.22          & 17.17          & {16.46} & 30.90  & \emph{30.82} & 0.735    \\
EduQG        & {29.19} & 21.69 & {18.03} & {16.76} & {33.18} & {34.36}  & {0.749} \\
\hline
% \emph{Our Proposals} &&&&&&&\\
Leaf+         & \emph{36.67}$^{(*)}$ & \emph{31.45}$^{(*)}$ & \emph{28.17}$^{(*)}$ & \textbf{24.26}$^{(*)}$ & \emph{41.65}$^{(*)}$ & \textbf{26.43}$^{(*)}$ & \emph{0.801}$^{(*)}$ \\
EduQG+        & \textbf{37.20}$^{(*)}$ & \textbf{33.86}$^{(*)}$ & \textbf{28.49}$^{(*)}$ & \emph{22.35}$^{(*)}$ & \textbf{43.04}$^{(*)}$ & 33.88$^{(*)}$ & \textbf{0.812}$^{(*)}$\\
\hline
\end{tabular}
\end{table}

% \begin{table}[] \centering \label{tab:resultsp}
% \caption{Comparison of predictive performance and linguistic quality between Leaf (baseline) and EduQG (our proposals). The best performance and second best performance is indicated in \textbf{bold} and \emph{italic} faces respectively.}
% \begin{tabular}{|c|c|c|c|c|c|c|}
% \hline
%                 % & \multicolumn{5}{c}{Predictive Performance}                          & \multicolumn{3}{c}{Linguistic Quality}       \\
% & {Model}           & BLEU-1   & BLEU-2   & BLEU-3 & BLEU-4  & F1-Score \\
% % &&&&&&&&\\

% \hline
% From & Leaf & 27.07          & 20.22          & 17.17          & {16.46} & 30.90    \\
% Table \ref{tab:results} &EduQG Large        & {29.19} & {21.69} & {18.03} & {16.76} & {33.18} \\ 
% \hline
% New & Leaf+        & \emph{36.67} & \emph{31.45} & \emph{28.17} & \textbf{24.26} & \emph{41.65} \\
% Proposals & EduQG+        & \textbf{37.20} & \textbf{33.86} & \textbf{28.49} & \emph{22.35} & \textbf{43.04} \\
% \hline
% \end{tabular}
% \label{tab:results}
% \end{table}

\newpage

\section{Discussion}

The results presented in section \ref{sec:results} provide sufficient information for us to answer the research questions pointed in section \ref{sec:rq}. 

\subsection{Ability of PLMs to Generate Educational Questions (RQ1)}

The results presented in Tables \ref{tab:human} and \ref{tab:results} together allow us to answer RQ1. It is seen from the linguistic quality metrics in Table \ref{tab:results} that the perplexity score obtained by all the trained models (both baseline and novel) is acceptable. That is, the perplexity scores obtained by the model-generated questions are much lower compared to the perplexity score of the SQuAD 1.1 questions that are human-generated. The language used in academic texts can be highly advanced and rich. This is reflected by the very low perplexity score and the high vocabulary diversity score of the SciQ questions in Table \ref{tab:human}. While the proposed models haven't achieved a perplexity score close to the SciQ questions, having a superior perplexity in comparison to 
% the human-generated questions in
the SQuAD 1.1 question shows that the generated questions inherit coherent language and human readability. The random examples presented in Table {\ref{fig:examples}} further reinforce this conclusion.   

\subsection{Effect of Pre-training with a Scientific Text Corpus (RQ2)}

% Table \ref{tab:results} provide observations on how the additional pre-training step that uses the S2ORC dataset impacts the quality of the answers generated. The 
Table \ref{tab:results} demonstrates that the novel models, EduQG Small and EduQG Large, surpass the baseline Leaf model in nearly all evaluation metrics for predicting educational questions in the SciQ test dataset. This improvement highlights the impact of additional pre-training on scientific text for generating educational questions. With all models fine-tuned using the same question generation dataset (the SQuAD dataset), the only intervention in the proposed models is during the pre-training phase as per \figurename{ \ref{fig:method}}.

The T5 language model, the foundational PLM in this experiment, is trained primarily on web-crawled data and Wikipedia articles \cite{raffel2020exploring}. However, this training corpus lacks scientific texts, leading to a weaker understanding of scientific knowledge and language. The improvement in predicting educational questions signifies that additional pre-training enhances the model's grasp of scientific knowledge and language, even without specific training on educational questions during fine-tuning.

Table \ref{tab:results} shows higher mean perplexity scores for EduQG models, though the difference is not statistically significant. \figurename{ \ref{fig:metrics} (iii)} indicates that the perplexity distribution between the two models is not statistically different. The observations in Table {\ref{fig:examples}} further illustrate that the EduQG model generates more educational and pedagogically sound questions, as seen in rows 3 and 4.   

\subsection{Impact of the Training Size on the Question Quality (RQ3)}

% In addition to depicting the usefulness of pre-training with scientific texts, t
The results in Table \ref{tab:results} further points out the performance difference between models \emph{EduQG Small} and \emph{EduQG Large} where the only difference is the size of pre-training data (green vs. yellow arrows in \figurename{ \ref{fig:method}}). The \emph{EduQG Large} model is superior in all evaluation metrics with the larger pre-training dataset of 23.2M data abstracts. The \emph{EduQG Small} model outperforms the baseline \emph{Leaf} model that uses fewer pre-training examples from the S2ORC dataset. This trend suggests that the increasing number of training examples used in the pre-training step leads to a better QG model. The increasing diversity values with the growing number of pre-training examples is also noticeable from Table \ref{tab:results}. The improvement of BLEU and F1-Scores with diversity indicates that the validity of questions is not harmed by the diversity of the vocabulary used by the model.

% caused by the number of scientific documents included in the pre-training dataset in figure \ref{fig:method}. The significantly smaller training dataset created from the S2ORC dataset (green path in \figurename{ \ref{fig:method}} leads to creating the \emph{EduQG Small} model. Table \ref{tab:results} show that the predicitve performance of the EduQG Small model is inferior to the EduQG Large model trained with the larger subset of 3.2M scientific abstracts although the EduQG Small model is significantly superior to the baseline leaf model. This observed trend suggests that the increasing number of training examples used in the pre-training step leads to a better question generation model ultimately, although all models are fine-tuned using the same SQuAD 1.1 dataset. The table further shows the growing diversity value stemming to the use of a richer vocabulary which the model learns from the abstracts in the S2ORC dataset. While the EduQG Large model, with more training data, leads to a larger diversity value, the improvement of the predictive performance (BLEU and F1 Scores) indicate that the model learns to use these new words coherently when generating questions.

\subsection{Effect of Fine-tuning Using Educational Questions (RQ4)}

% Now that RQs 1-3 has established the positive effect of pre-training the PLM with scientific text, the final RQ aims to explore the impact of further fine-tuning the model with a more task focused educational question datset.
The experimental setups of RQ 2 and 3 use a \emph{zero-shot} evaluation where no observations from the SciQ dataset are used during the training phase. On the contrary, the experiments relating to RQ4 ( $\cdot \ +$ models in \figurename{ \ref{fig:method}}) use the training data from the SciQ dataset that allows the newly proposed models, \emph{Leaf+} and \emph{EduQG+} to learn from educational question examples. Table \ref{tab:resultsp} indicates that the additional fine-tuning significantly improves the predictive accuracy. It is noteworthy that fine-tuning is also improving the perplexity score of the generated questions which was absent in the previous experiments. We can see that the new models are outperforming the baselines. This improvement attributes to the low perplexity score of SciQ questions as per Table \ref{tab:human} that are exposed to the model during training.

% In the previous experiments, while we managed to improve predictive quality of the educational questions generated, the baselines dominated the preplexity score. 
% However, when fine-tuned with educational questions, we can see that the new models outperform baselines. This also attributes to the low perplexity score of the human-generated questions in the SciQ dataset as per table \ref{tab:human}. When the model gets exposed to generating questions similar to the quesitons in SciQ dataset during training, it learns to use similar language leading to lower perplexity.

\subsection{Opportunities}

The examples in Table {\ref{fig:examples}} with all the above results indicate that  educational QG systems are very close to becoming part of human-facing technology-enhanced learning systems (Such as X5Learn that leverages Open Educational Resources \cite{x5learn}). Many works in the past have shown how zero-shot question generation is operationally feasible using very large language models gated behind an API from a large corporation (Model-as-a-Service architecture) \cite{wang2022towards}. However, our result contributes to this topic as we introduce methods to enhance openly-available PLMs (in our case, T5) to support educational QG. We intentionally use the \emph{T5-Small} model that has 60M parameters in comparison models such as GPT-3 XL that has 1.3B parameters \cite{brown2020language} to show that relatively small models can be trained with domestic hardware to create SOTA educational QG capabilities. Our method also gives the stakeholder full control and ownership, a critical feature for quality assurance of the downstream educational systems that rely on this model (contrary to having no control over a third party that can change their model over time). This work also informs the educational data mining community that domain-specific data can be used with language models to harness them to specific educational use cases (e.g. extend to other domains, different question types that support diverse pedagogy etc.). While the proposed systems are not perfect, the quality of AI-generated questions indicates that a teacher or an educator can re-purpose these questions with minimum effort and time. Human-in-the-loop systems can be built to support educators while their corrections will harvest more training data to improve the models over time. Educational questions can be generated at scale using the proposed model both for existing and newly created learning resources, adding more testing opportunities for learners/teachers to use when needed.  

We see our work being foundational to building a series of tools that can support educators with scalable/personalised assessments. Ultimately, we have the opportunity to improve these models to the point where an intelligent tutor can rely on them to create on-demand questions to verify a learner's knowledge state with no human intervention.

\subsection{Limitations}
% While automatic QG models can revolutionise how technology enhanced education and personalise education will evolve, 
We need to be cautious to avoid the obvious pitfalls of such automatic systems. Intelligent QG models we build tend to exhibit the patterns in the data that we feed them. We need to be mindful that we take rigorous steps to validate the datasets to be ethically and pedagogically sound. Putting emphasis on quality assurance of the training data will help us to build ethical, unbiased QG models that can benefit all learners equally. 

Many intelligent learning systems exploit learner engagement signals to determine what characteristics of the system should sharpen and weaken \cite{truelearn}. In the context of question generation, it is important to distinguish between \emph{bad} questions vs. \emph{difficult} questions as the latter, although demanding, may positively impact a learner while the former will only hinder and diminish learning gains. The AI-generated questions should allow users to improve their learning gains over time. 

Another gap in this work is the lack of human evaluation of the AI-generated questions. While offline evaluation on labelled datasets is useful, having teachers and learners evaluate and contrast between human vs. AI-generated questions will provide much more insightful findings that can improve this line of research in the future. Our subsequent work will focus on this aspect.  

% \begin{table}[] \centering \
% % small
% \begin{tabular}{c|ccccc|ccc}
% \hline
%                 & \multicolumn{5}{c}{Predictive Performance}                          & \multicolumn{3}{c}{Linguistic Quality}       \\
% Model           & BLEU-1 $\uparrow$     & BLEU-2 $\uparrow$     & BLEU-3 $\uparrow$     & BLEU-4   $\uparrow$   & F1-Score  $\uparrow$       & Perplexity $\downarrow$       & Diversity $\uparrow$       & Grammar Errors $\downarrow$       \\
% % &&&&&&&&\\

% \hline
% Leaf (Baseline) & 27.07          & 20.22          & 17.17          & {16.46} & 30.90          & \textbf{30.82} & 0.735          & \textbf{0.102} \\
% EduQG (Ours)        & \textbf{29.19} & \textbf{21.69} & \textbf{18.03} & \textbf{16.76} & \textbf{33.18} & 34.36          & \textbf{0.749} & 0.122  \\       
% \hline
% \end{tabular}
% \caption{Comparison of predictive performance and linguistic quality between Leaf (baseline) and EduQG (our proposal). The superior performance is indicated in \textbf{bold} face.}
% \label{results}
% \end{table}

\section{Conclusion}
This work demonstrates the operational feasibility of adapting pre-trained language models for educational question generation. Specifically, we argue that a relatively small language model manageable with domestic hardware can be further trained and harnessed with low computational costs and produce a humanly-acceptable educational question generation model. We validate that a PLM fine-tuned with question generation data can generate questions that are linguistically valid and humanl-like. We show that the quality of the educational questions generated can be significantly improved by pre-training using domain-specific corpora alone. We use a corpus of scientific abstracts to empirically demonstrate this while we point out the relationship between the prediction quality and the amount of data. Going further, we improve the model's question generation capabilities significantly by further fine-tuning it using a domain-specific question dataset, indicating fine-tuning can be used to further improve the model. 

A few promising steps remain to take this work to the future. Validating the generalisability of our approach to other PLMs such as GPT \cite{brown2020language} and extending evaluation to human experts \cite{bhat2022towards,wang2022towards} are the immediate next steps. Establishing methods to audit the ethical and pedagogical value of training datasets will improve the use of the downstream QG models. Identifying systematic approaches (e.g. using curriculum learning) to identify the most useful training examples would allow us to make QG models significantly better with less number of training examples leading to computational cost savings. Finally, formalising concepts such as question difficulty, and value for learning will allow us to evaluate the quality of generated questions much more pragmatically. 

\subsubsection{Acknowledgements} 
This work is also partially supported by the European Commission-funded project "Humane AI: Toward AI Systems That Augment and Empower Humans by Understanding Us, our Society and the World Around Us" (grant 820437), EU Erasmus+ project 621586-EPP-1-2020-1-NO-EPPKA2-KA and the EPSRC Fellowship "Task Based Information Retrieval" (grant EP/P024289/1). This  research  is  conducted  as  part  of  the  X5GON  project (\url{www.x5gon.org}) funded by the EU’s Horizon 2020 grant No 761758.

\bibliographystyle{splncs04}
\bibliography{mybib}




% LNCS Homepage, \url{http://www.springer.com/lncs}. Last accessed 4
% \end{thebibliography}
\end{document}